\documentclass{article}

\usepackage[preprint]{neurips_2026}

\usepackage[utf8]{inputenc} 
\usepackage[T1]{fontenc}    
\usepackage{hyperref}       
\usepackage{url}            
\usepackage{booktabs}       
\usepackage{amsfonts}       
\usepackage{nicefrac}       
\usepackage{microtype}      

\usepackage{graphicx}
\usepackage{subcaption}

\usepackage{amsmath}
\usepackage{amssymb}
\usepackage{mathtools}
\usepackage{amsthm}
\usepackage{algorithm}
\usepackage{algorithmic}

\usepackage[capitalize,noabbrev]{cleveref}

\theoremstyle{plain}
\newtheorem{theorem}{Theorem}[section]

\newtheorem{lemma}[theorem]{Lemma}

\theoremstyle{definition}

\newtheorem{assumption}[theorem]{Assumption}
\theoremstyle{remark}

\usepackage{fancybox}
\usepackage[table]{xcolor}
\usepackage{longtable}
\usepackage{array}
\usepackage{colortbl}
\usepackage{tcolorbox}
\tcbuselibrary{breakable,skins}

\renewcommand{\textcolor}[2]{#2}

\newcommand{\codeavailability}{}

\newcommand{\scorecell}[1]{%
  \begingroup
  \ifnum#1<3 \cellcolor{red!18}\textbf{#1}\else
  \ifnum#1<6 \cellcolor{orange!18}\textbf{#1}\else
  \ifnum#1<8 \cellcolor{yellow!20}\textbf{#1}\else
  \cellcolor{green!18}\textbf{#1}%
  \fi\fi\fi
  \endgroup
}

\title{Gradient-Aligned Pair Selection for Personalized Preference Optimization}

\author{%
  Ruoming Jin\thanks{Corresponding author.} \\
  Kent State University \\
  \texttt{rjin1@kent.edu}
  \And
  Xinyu Li \\
  Kent State University \\
  \texttt{xli74@kent.edu}
  \And
  Hao Zhou \\
  Kent State University \\
  \texttt{hzhou6@kent.edu}
  \AND
  Jianfeng Zhu \\
  Kent State University \\
  \texttt{jzhu10@kent.edu}
  \And
  Ruixin Guo \\
  Kent State University \\
  \texttt{rguo5@kent.edu}
  \And
  Feodor Dragan \\
  Kent State University \\
  \texttt{fdragan@kent.edu}
  \AND
  Lei Xu \\
  Kent State University \\
  \texttt{lxu12@kent.edu}
  \And
  Haixun Wang \\
  EvenUp, USA \\
  \texttt{haixun.wang@evenup.ai}
  \And
  Yang Zhou \\
  Auburn University \\
  \texttt{yangzhou@auburn.edu}
}

\begin{document}

\maketitle

\begin{abstract}
Personalizing large language models (LLMs) requires aligning generation behavior with user-specific preferences rather than aggregate quality. While Direct Preference Optimization (DPO) provides a stable framework for preference learning, its effectiveness in personalized settings critically depends on how preference pairs are selected. Existing approaches typically rely on heuristic criteria, such as likelihood-based extremes, which decouple optimization from explicit user utility and can lead to degraded personalization.

We formalize personalized preference learning as a geometry-aligned optimization problem by analyzing the first-order interaction between gradients of expected user utility and DPO update directions. Our analysis reveals that, under off-policy sampling, the DPO update transitions from a purely error-corrective signal to a reinforcement-like update when preference margins are directionally aligned with utility gradients. This perspective exposes pair selection as a geometric decision that governs whether preference optimization advances or hinders personalization.

Motivated by this insight, we propose GAP-DPO (Geometry-Aligned Preference DPO), an iterative algorithm that performs utility-aware, geometry-aligned pair selection while controlling distribution shift via epoch-wise regeneration. Experiments on personalized text generation benchmarks show that GAP-DPO consistently improves stylistic fidelity, preference alignment, and generation quality compared to standard DPO variants. Together, our results establish gradient alignment as a unifying principle for personalized preference optimization and demonstrate that pair selection is an intrinsic component of the optimization geometry rather than a heuristic preprocessing step.

\end{abstract}

\section{Introduction}

Large Language Models (LLMs) have achieved remarkable success through alignment with human
preferences~\cite{zhao2023survey,xu2024contrastive, abbasiantaeb2024let, ferrag2025llm}, producing outputs that are broadly helpful, fluent, and safe. However, most
existing alignment methods implicitly adopt a \emph{one-size-fits-all} notion of quality,
optimizing for aggregate preferences across users. In practice, user satisfaction depends
critically on individual preferences for tone, verbosity, formality, and rhetorical
structure ~\cite{fan2024survey,zhao2023survey,wu2025webwalker,chu2025llm}. A legal professional may favor precise and formal language, while a creative
writer values vivid and unconventional expression. Standard alignment pipelines average
over these distinctions, systematically erasing user identity in favor of a generic
assistant style.

Personalized alignment reframes the learning objective from asking \emph{which response is
better on average} to asking \emph{which response is better for a specific user}. When
different users respond to the same prompt, their variations encode rich stylistic
signals—spanning lexical choice, syntactic structure, and rhetorical form—that naturally
induce preference comparisons~\cite{fan2024survey,zhao2023survey,wu2025webwalker,chu2025llm}. Leveraging these comparisons is central to learning
personalized generation behavior.

\paragraph{The Landscape of Personalized LLMs.}
Existing approaches to personalized LLMs broadly fall into three categories.
\emph{Prompt-based methods}~\cite{salemi2023lamp, liu2021pre, wang2023learning, kang2023llms, qiu2025measuring}, often combined with retrieval-augmented generation (RAG),
inject user history or profile information directly into the prompt. While lightweight and
interpretable, these approaches are constrained by context length and rely heavily on
explicit user signals.
\emph{Parameter-efficient fine-tuning (PEFT)} methods, such as user-specific LoRA adapters,
offer finer-grained control but incur storage overhead and are prone to overfitting in
low-data regimes ~\cite{tan2024personalized, zhang2024personalized1, zhao2025nextquill,tan2024democratizing, bu2025personalized}.
\emph{Personalized reward modeling} ~\cite{bose2025lore,shenfeld2025language,ryan2025synthesizeme,seo2026p,poddar2024personalizing,nam2025learning} seeks to distill user preferences into a learned reward
function, which then guides policy optimization via PPO or DPO~\cite{ziegler2019finetuning,rafailov2023direct}. Although expressive, these
methods introduce additional instability and hinge on the quality of a learned reward
model that is difficult to estimate reliably for individual users.

A common assumption underlying many of these frameworks is access to explicit preference
annotations or ranked pairs. In practice, however, such labels are rare for open-ended text
generation. To bypass this limitation, recent work~\cite{bu2025personalized,seo2026p,chen2024bootstrapping} construct preference signals
automatically, often adopting the \emph{implicit reward} structure of Direct Preference
Optimization (DPO)~\cite{rafailov2023direct}:
\begin{equation}
r_\theta(x,y)
=
\log \pi_\theta(y \mid x)
-
\log \pi_{\mathrm{ref}}(y \mid x).
\end{equation}
DPO optimizes a Bradley--Terry-style objective over inferred preference pairs, avoiding the
need for an explicit reward model and enabling stable training.

\paragraph{The Pair Selection Problem.}
When explicit preference labels are unavailable, the effectiveness of DPO hinges critically
on \emph{how preference pairs are constructed}. One line of work exploits
\emph{inter-user contrastive learning}, where responses from distant users serve as
negatives; for example, \textsc{P-Check}~\cite{seo2026p} dynamically constructs checklists by contrasting
responses across users to uncover stylistic utility. Other approaches rely on
bootstrapping: generating multiple candidates and selecting pairs with the highest and
lowest implicit rewards (i.e., likelihood ratios) as proxies for preference~\cite{bu2025personalized,chen2024bootstrapping}.

Despite their empirical success, these strategies are largely heuristic. Selecting the
lowest-likelihood response often produces pairs that are trivially separable and provide
little useful learning signal, while harder, high-margin pairs may offer greater refinement
but lack a principled selection rule. More importantly, likelihood-based criteria are
agnostic to \emph{user-specific utility}~\cite{fan2024survey,zhao2023survey,wu2025webwalker,chu2025llm}. As a result, the implicit rewards driving
optimization may be decoupled from the true objectives of personalization, particularly for
stylistic attributes that are difficult to encode probabilistically.

This issue is especially pronounced when preserving \emph{stylistic fidelity}. External
evaluators, such as LLM-based style checkers or persona consistency metrics~\cite{salemi2023lamp,kumar2024longlamp}, can measure these
qualities, but their signals do not naturally align with likelihood-ratio objectives.
Consequently, optimizing inferred preferences can improve the surrogate DPO loss while
degrading the user’s perceived quality~\cite{bu2025personalized}.

\paragraph{Personalization Divergence and Gradient Alignment.}
We identify a fundamental failure mode in personalized preference learning:
\emph{gradient misalignment}. In standard DPO, each preference pair induces a contrastive
gradient that increases the probability of the ``preferred'' response and decreases that
of the alternative. However, this update direction does not necessarily align with the
gradient of the user’s true utility. We refer to this phenomenon as
\emph{personalization divergence}, a regime in which preference optimization progresses,
yet personalized generation quality deteriorates.

This observation motivates a central question that has been largely overlooked:
\begin{quote}
\emph{Given a set of candidate responses, which preference pairs should be optimized, and
why?}
\end{quote}

\paragraph{Our Approach.}
We address this question by formulating personalized preference learning as a
\emph{geometry-aware optimization problem}. We analyze the first-order interaction between
the gradient of expected user utility and the DPO update direction, deriving a criterion
that determines whether optimizing a given preference pair advances or hinders
personalization.

Based on this analysis, we propose \textbf{GAP-DPO} (Geometry-Aligned Preference DPO), an
iterative off-policy algorithm that performs \emph{utility-aware, geometry-aligned pair
selection}. Rather than treating pair selection as a heuristic preprocessing step,
GAP-DPO selects contrastive pairs whose induced DPO updates are directionally aligned with
maximizing expected user utility. To ensure stability under off-policy training, the
algorithm regenerates candidate pools epoch-wise from a policy close to the current
iterate, controlling distribution shift and preserving the validity of the geometric
approximation.

Our contributions are threefolds:
\begin{itemize}
\item We diagnose \emph{personalization divergence} as a consequence of gradient
misalignment in contrastive preference learning.
\item We derive a principled, geometry-aware pair selection criterion based on first-order
utility improvement.
\item We demonstrate empirically that GAP-DPO significantly improves stylistic fidelity
and preference alignment across diverse personalized generation benchmarks.
\end{itemize}

Together, these results show that effective personalization requires not only preference
data, but principled control over \emph{which} preferences are optimized, transforming
personalized alignment from a heuristic data problem~\cite{bu2025personalized,seo2026p,chen2024bootstrapping} into a geometry-driven optimization
problem.

\section{Preliminaries and Problem Definition}

\begin{table}[t]
\centering
\small
\begin{tabular}{ll}
\toprule
\textbf{Symbol} & \textbf{Description} \\
\midrule
$x \in \mathcal{X}$ & Input prompt (e.g., instruction or query) \\
$y \in \mathcal{Y}$ & Textual response generated by the LLM \\
$u$ & User index \\
$\textcolor{blue}{h_u}$ & \textcolor{blue}{user-specific context or
latent preferences} \\
$\pi_\theta(y \mid x; \textcolor{blue}{h_u})$ & Target policy parameterized by $\theta$ \\
$\pi_{\mathrm{ref}}(y \mid x; \textcolor{blue}{h_u})$, $\pi_0$ & Fixed reference (behavior) policy \\
$p_\theta(y \mid x)$ & Policy probability $\pi_\theta(y \mid x)$ \\
\textcolor{red}{$\Lambda_\theta(y \mid x)$} & Log-likelihood ratio $\log \frac{\pi_\theta(y \mid x)}{\pi_0(y \mid x)}$ \\
$w_\theta(y \mid x)$ & Importance weight $\frac{\pi_\theta(y \mid x)}{\pi_0(y \mid x)}$ \\
$g(y)$ & Score function $\nabla_\theta \log \pi_\theta(y \mid x)$ \\
$y^+, y^-$ & Preferred and non-preferred responses \\
$g^\pm$ & Shorthand for $g(y^\pm)$ \\
$w^\pm$ & Shorthand for $w_\theta(y^\pm \mid x)$ \\
\bottomrule
\end{tabular}
\caption{Summary of notation used throughout the paper. Conditioning on $(x, \textcolor{blue}{h_u})$ is omitted when unambiguous.}
\label{tab:notation}
\end{table}

Table~\ref{tab:notation} summarizes the key notations used in the paper. 
In this study, we focus on preference learning methods, with particular emphasis on
Direct Preference Optimization (DPO)~\cite{rafailov2023direct} and its variants~\cite{azar2024general,achiam2017constrained,ethayarajh2024kto, hong2024orpo}.
Empirically, we find that DPO exhibits superior stability and performance in
personalized preference learning settings (Appendix~\ref{app:DPO-variants} and  Table~\ref{tab:dpo_variants}).
Accordingly, our analysis and algorithmic development center on DPO, although the
proposed framework is not DPO-specific and can be extended to other preference
optimization objectives.

\paragraph{DPO Objective.}
We define the implicit reward induced by policy deviation as
\begin{equation}
r_\theta(x,y)
=
\log \pi_\theta(y \mid x)
-
\log \pi_{\mathrm{ref}}(y \mid x).
\end{equation}
The Direct Preference Optimization (DPO) ~\cite{rafailov2023direct} objective is given by
\begin{equation}
\mathcal{L}_{\mathrm{DPO}}
=
-\mathbb{E}
\left[
\log \sigma
\left(
\beta \big(
r_\theta(x,y^+) - r_\theta(x,y^-)
\big)
\right)
\right], \label{eq:3}
\end{equation}
where \(\sigma(\cdot)\) denotes the sigmoid function and \(\beta > 0\) controls the
sharpness of preference separation.
This objective corresponds to a Bradley--Terry (or Thurstone) pairwise comparison model,
in which \(\sigma(\beta(\cdot))\) represents the probability that \(y^+\) is preferred
over \(y^-\) given latent utility \(r_\theta(x,\cdot)\).


\paragraph{External Utilities and Evaluation Objective.}
Let \(q(y)\) denote a scalar utility or evaluation score associated with response \(y\),
such as ROUGE, BLEU, human preference scores, or outputs from an automated evaluator~\cite{tan2025personabench,salemi2023lamp,kumar2024longlamp}.
In contrast to classical reinforcement learning, we do \emph{not} seek to learn or adapt
this utility function.
Instead, \(q(\cdot)\) is treated as a fixed, external evaluation criterion.

Our ultimate objective is to maximize the expected personalized utility
\begin{equation}
F
=
\mathbb{E}_{u}
\mathbb{E}_{x}
\mathbb{E}_{y \sim \pi_\theta(\cdot \mid x; \textcolor{blue}{h_u})}
\big[
q(y \mid x; \textcolor{blue}{h_u})
\big].
\end{equation}

\paragraph{The Pair Selection Problem.}
Classical reinforcement learning methods, such as PPO, typically assume access to
explicit pairwise rankings or preference labels, either directly or via a learned
reward model.
In personalized text generation, however, such ranked pairs are rarely observed.
Most datasets instead contain only a single response per prompt, often already
consumed during supervised fine-tuning~\cite{bu2025personalized,seo2026p}.

Motivated by recent work on synthetic preference construction (e.g., DICE~\cite{chen2024bootstrapping}, CoPE~\cite{bu2025personalized}),
we consider the following pair selection procedure.
For a given prompt \(x\) and user \(u\), we draw a pool of candidate responses
\[
\{y_1, \dots, y_n\}
\sim
\pi_0(\cdot \mid x; \textcolor{blue}{h_u}),
\]
from which a positive--negative pair \((y^+, y^-)\) is selected according to a
specified criterion.
Here, \(y^+\) denotes the preferred (``winner'') response and \(y^-\) the less
preferred (``loser'') response.

Existing approaches typically rely on \emph{implicit rewards}, such as log-likelihood
ratios, selecting the most likely response as \(y^+\) and the least likely as \(y^-\)~\cite{chen2024bootstrapping,bu2025personalized,seo2026p}.
However, such criteria are agnostic to the true downstream evaluation objective
\(q(\cdot)\).
As a result, preference optimization may improve the surrogate objective while
degrading actual personalized generation quality, a phenomenon we refer to as
\emph{personalization divergence}.
This effect is further amplified when preference signals are noisy, derived from
cross-user comparisons, or generated automatically, where superficial stylistic cues
can dominate the loss.
\vspace{-3mm}
\paragraph{Problem Statement.}
The central problem we study is therefore:
\begin{quote}
\emph{How should preference pairs be selected so that optimization under a DPO-style
objective provably improves the true evaluation objective \(F\)?}
\end{quote}

\section{Approach}

We emphasize two key distinctions relative to standard on-policy preference
optimization.
First, candidate responses are sampled from a \emph{frozen reference policy}
\(\pi_{\mathrm{ref}} \equiv \pi_0\), rather than the current policy \(\pi_\theta\).
Second, improvement is evaluated through the \emph{DPO update direction}~\cite{rafailov2023direct},
allowing us to analyze first-order changes in the true expected external utility.

\paragraph{Global and local objectives.}
We define the overall population-level objective as
\begin{equation}
\label{eq:global_objective}
F(\theta)
=
\mathbb{E}_{u}
\mathbb{E}_{x}
\mathbb{E}_{y \sim \pi_\theta(\cdot \mid x; \textcolor{blue}{h_u})}
\big[ q(y \mid x; \textcolor{blue}{h_u}) \big].
\end{equation}
\textcolor{red}{For a fixed user \(u\) and prompt \(x\), we define the conditional expected utility
\begin{equation}
\label{eq:local_objective}
f(\theta; u, x)
:=
\mathbb{E}_{y \sim \pi_\theta(\cdot \mid x; \textcolor{blue}{h_u})}
\big[ q(y \mid x; \textcolor{blue}{h_u}) \big].
\end{equation}
Since both pair selection and DPO updates are performed independently for each
\((u,x)\), analyzing improvements of \(f(\theta; u, x)\) suffices to characterize
optimization of the global objective \(F(\theta)\).
Accordingly, in the remainder of this section we fix \((u,x)\) and, with slight
abuse of notation, write \(f(\theta)\) for brevity}.

Unless stated otherwise, all probabilities are implicitly conditioned on
\((x, \textcolor{blue}{h_u})\).

\paragraph{Personalized Objective and Off-Policy Evaluation.}
Our goal is to maximize the expected external utility
\(q(y \mid x; \textcolor{blue}{h_u})\) under a personalized target policy \(\pi_\theta\),
where \(\textcolor{blue}{h_u}\) encodes user-specific context or latent preferences. \textcolor{blue}{In practice, $h_u$ is usually obtained by retrieving the top-$k$ most similar interactions from the user's history and directly including them in the prompt}.
For a fixed user \(u\) and prompt \(x\), we define
\begin{equation}
\label{eq:global_f}
f(\theta)
=
\mathbb{E}_{y \sim \pi_\theta(\cdot \mid x; \textcolor{blue}{h_u})}
\big[ q(y \mid x; \textcolor{blue}{h_u}) \big].
\end{equation}

In our setting, candidate responses are generated by a fixed
\emph{behavior policy} \(\pi_0 \equiv \pi_{\mathrm{ref}}\),
typically obtained via supervised fine-tuning.
The target policy \(\pi_\theta\) is optimized using these samples but is treated
as fixed within each optimization step; we do not perform online or iterative
resampling.
Accordingly, evaluation of \eqref{eq:global_f} is carried out off-policy via
importance sampling~\cite{schulman2017proximalpolicyoptimizationalgorithms}:
\begin{equation}
\label{eq:is_f}
\begin{aligned}
f(\theta)
&=
\mathbb{E}_{y \sim \pi_0(\cdot \mid x; \textcolor{blue}{h_u})}
\left[
q(y \mid x; \textcolor{blue}{h_u})
\frac{\pi_\theta(y \mid x; \textcolor{blue}{h_u})}{\pi_0(y \mid x; \textcolor{blue}{h_u})}
\right] \\
&=
\mathbb{E}_{y \sim \pi_0}
\big[
q(y)\, w_\theta(y)
\big],
\end{aligned}
\end{equation}
where
\(
w_\theta(y)
=
\pi_\theta(y \mid x; \textcolor{blue}{h_u})/\pi_0(y \mid x; \textcolor{blue}{h_u})
\)
denotes the importance weight.

\paragraph{Utility Gradient and Pairwise Estimation.}
Differentiating \eqref{eq:is_f} and using
\(
\nabla_\theta \pi_\theta
=
\pi_\theta \nabla_\theta \log \pi_\theta
\),
the gradient of the local objective is
\begin{equation}
\label{eq:full_grad}
\nabla_\theta f
=
\mathbb{E}_{y \sim \pi_0}
\big[
w_\theta(y)\, q(y)\, \nabla_\theta \log \pi_\theta(y)
\big].
\end{equation}
DPO-style optimization operates on response pairs \((y^+, y^-)\).
For a selected pair, we consider the stochastic pairwise estimator
\begin{equation}
\label{eq:pairwise_grad}
\widehat{\nabla}_\theta f
\approx
w^+ q^+ g^+ + w^- q^- g^-,
\end{equation}
where
\(g^\pm = \nabla_\theta \log \pi_\theta(y^\pm)\).
In the on-policy case (\(\pi_\theta = \pi_0\)), the importance weights satisfy
\(w^\pm = 1\), recovering a standard REINFORCE-style estimator~\cite{williams1992simple}.
In the off-policy setting, \(w^\pm\) correct for the discrepancy between
\(\pi_\theta\) and \(\pi_0\).

\paragraph{DPO Update Direction.}
For a response pair \((y^+, y^-)\), define the log-probability gaps
\begin{align}
\Delta_\theta(x, y^+, y^-)
&=
\log \pi_\theta(y^+ \mid x; \textcolor{blue}{h_u})
-
\log \pi_\theta(y^- \mid x; \textcolor{blue}{h_u}), \\
\Delta_{\mathrm{ref}}(x, y^+, y^-)
&=
\log \pi_{\mathrm{ref}}(y^+ \mid x; \textcolor{blue}{h_u})
-
\log \pi_{\mathrm{ref}}(y^- \mid x; \textcolor{blue}{h_u}).
\end{align}
Using this notation, Eq.~\eqref{eq:3} can be written as
\begin{equation}
\mathcal{L}_{\mathrm{DPO}} =
-\mathbb{E}_{(x,y^+,y^-)}
\left[
\log \sigma
\left(
\beta
\big[
\Delta_\theta(x,y^+,y^-)
-
\Delta_{\mathrm{ref}}(x,y^+,y^-)
\big]
\right)
\right],
\end{equation}

Then, the DPO update direction (gradient) is given by
\begin{equation}
\label{eq:dpo_direction}
d_{\mathrm{DPO}}
=
\beta \big( 1 - \sigma(z_\theta) \big)
\big( g^+ - g^- \big),
\qquad
z_\theta
=
\beta \big(
\Delta_\theta - \Delta_{\mathrm{ref}}
\big).
\end{equation}

\paragraph{Utility Improvement and Alignment.}
We define the first-order improvement in expected utility induced by a DPO step as
\[
\delta f
:=
\widehat{\nabla}_\theta f \cdot d_{\mathrm{DPO}}.
\]
Substituting \eqref{eq:pairwise_grad} and \eqref{eq:dpo_direction} yields 
\begin{equation}
\label{eq:delta_f_full}
\boxed{
\begin{aligned}
\delta f
= & \beta \big( 1 - \sigma(z_\theta) \big)
\Big[
w^+ q^+
\big( \|g^+\|^2 - \langle g^+, g^- \rangle \big)
\\
&\qquad\quad
+
w^- q^-
\big( \langle g^+, g^- \rangle - \|g^-\|^2 \big)
\Big]
\end{aligned}
}
\end{equation}


Equation~\eqref{eq:delta_f_full} precisely characterizes how the utility improvement depends on
(i) the external utility gap,
(ii) off-policy importance weights, and
(iii) the geometric interaction between score gradients.
However, directly evaluating this quantity requires computing and storing
per-sample gradients and their inner products, which is prohibitively expensive
for large language models and incompatible with efficient batched training.
Moreover, such exact gradient-level information is unnecessary for guiding
pair selection at scale.

\paragraph{Simplification via Approximate Orthogonality.}
To obtain a tractable and interpretable criterion, we consider a high-dimensional
regime in which score gradients associated with distinct samples are weakly
correlated. Empirically, for large models, especially under parameter-efficient
fine-tuning, gradients corresponding to different generations of the same prompt
exhibit low cosine similarity, while their norms concentrate around a common
scale. Formally, we adopt the approximations
\[
\langle g^+, g^- \rangle \approx 0,
\qquad
\|g^+\|^2 \approx \|g^-\|^2 \approx C,
\]
where $C$ is a constant that depends on the local curvature of the model~\cite{ma2025gradientimbalancedirectpreference,yuan2025commonpitfallmarginbasedlanguage}.

Under these assumptions, the exact expression in
\eqref{eq:delta_f_full} simplifies to
\begin{equation}
\label{eq:delta_f_simple}
\boxed{
\delta f
\approx
C \beta \big( 1 - \sigma(z_\theta) \big)
\big(
w^+ q^+ - w^- q^-
\big).
}
\end{equation}

This simplified form reveals that, up to a positive scaling factor, the sign of
the utility improvement is determined solely by the importance-weighted utility
gap between the preferred and non-preferred responses.
Crucially, this criterion depends only on quantities that are inexpensive to
compute, external utilities, policy logits, and importance ratios, making it
suitable for large-scale training and iterative pair selection.

While the approximation above is used to motivate an efficient selection rule,
our theoretical results in Appendix~\ref{app:monotonicity} relax the strict
orthogonality assumption and show that the same qualitative conclusion holds
under bounded gradient correlation, ensuring that the simplified criterion
remains directionally consistent with true utility ascent.



\paragraph{Special Case (Utility Gap): Initialization (\(\pi_\theta \approx \pi_{\mathrm{ref}}\)).}
At initialization, we have \(\pi_\theta \approx \pi_{\mathrm{ref}}\), implying
\(w^+ \approx w^- \approx 1\) and \(z_\theta \approx 0\).
Consequently,
\begin{equation}
\label{eq:delta_f_init}
\boxed{
\delta f_{\mathrm{init}}
\approx
\frac{C \beta}{2}
\big(
q^+ - q^-
\big).
}
\end{equation}
At this stage, the DPO update is driven purely by the raw external utility
difference between the preferred and non-preferred responses, independent of
model confidence.
We refer to the criterion in Eq.~\eqref{eq:delta_f_init} as the \emph{Utility Gap (UG)}.

\paragraph{General Case: Proxy-Guided Pair Selection.}
As training progresses, preference pairs may be selected using a
\emph{proxy logit} \(\hat{z}\) computed from a past or auxiliary model
\(\pi_{\mathrm{past}}\).
This proxy is used only for pair selection; the update itself is applied
to the current target policy \(\pi_\theta\).

Replacing the true DPO logit \(z_\theta\) by \(\hat{z}\) in the update
direction, the first-order expected utility improvement takes the form
\begin{equation}
\label{eq:delta_f_proxy}
\delta f
\;\approx\;
C \beta \big( 1 - \sigma(\hat{z}) \big)
\big(
w^+ q^+ - w^- q^-
\big),
\end{equation}
where \(w^\pm\) and \(q^\pm\) are evaluated under the current model
\(\pi_\theta\).

Under a mild \emph{local stationarity} assumption—namely that
\(\pi_\theta\) does not drift significantly from \(\pi_{\mathrm{past}}\)
over the course of pair selection—we may interpret \(\hat{z}\) as a
reliable proxy for the true margin between \(y^+\) and \(y^-\).
In this regime, \(\sigma(-\hat{z})\) acts as a smooth gating factor that
downweights low-margin or ambiguous pairs and emphasizes confident
preferences.

Equation~\eqref{eq:delta_f_proxy} reveals that DPO-style updates optimize
expected utility through a \emph{gated difference of utilities}.
The update magnitude is jointly controlled by: 1)  the external utility contrast \(q^+ - q^-\); 2)  the importance weights \(w^\pm\), correcting for off-policy sampling;
3) and the confidence gate \(1 - \sigma(\hat{z})\), which modulates how
aggressively a pair is reinforced.

\paragraph{Dropping Importance Weights (Error-Gated Utility Gap).}
In principle, the importance weights $w^\pm$ correct for the mismatch between
the policy used to generate candidate responses and the current target policy.
However, in many practical preference optimization settings, candidate pools are
generated from a policy that remains locally close to the current iterate over
the time scale of pair selection. Under this \emph{local stationarity} regime,
the log-density ratio between the two policies is uniformly small on the sampled
support, implying
\[
w(y) = 1 + \mathcal{O}(\epsilon)
\quad \text{for all } y \in \mathcal{Y}_{\mathrm{gen}}.
\]

When this holds, the importance weights contribute only higher-order corrections
and can be safely dropped to obtain a simpler and lower-variance approximation.
Substituting $w^+ \approx w^- \approx 1$ into
Equation~\eqref{eq:delta_f_proxy} yields
\begin{equation}
\label{eq:delta_f_proxy_simplified}
\boxed{
\delta f
\;\approx\;
C \beta \big( 1 - \sigma(\hat{z}) \big)\,(q^+ - q^-).
}
\end{equation}
We refer to this selection criterion as the \emph{Error-Gated Utility Gap (EG-UG)}.

This form preserves the essential geometry of the update: expected utility
improvement is governed by a \emph{utility gap} that is modulated by an
\emph{error-sensitive gate}. The factor $1-\sigma(\hat z)=\sigma(-\hat z)$
emphasizes pairs for which the model is uncertain or misranked, focusing
optimization on correcting errors while suppressing updates for already
confident preferences. Crucially, this approximation avoids explicit importance
weight estimation, which is often noisy and computationally expensive in
large-scale language model training. It is also worth noting that
Eq.~\eqref{eq:delta_f_proxy_simplified} reduces to vanilla DPO when
$q^{+}=1$ and $q^{-}=-1$. Equation~\eqref{eq:delta_f_proxy} is also
consistent with reinforcement learning with verifiable rewards, such as in
mathematics and coding tasks where the reward function is verifiable.

Although importance weights may be dropped under local stationarity, the sigmoid
gate $\sigma(-\hat z)$ remains essential: it enforces confidence-aware update
magnitudes and ensures that learning is concentrated near the decision boundary,
where corrective signals are most informative.

\subsection{Algorithm: Off-Policy DPO with Utility-Aware Pair Selection}
\label{sec:algorithm}

We now instantiate the geometric alignment analysis into a practical training
procedure. While our theoretical framework naturally admits an iterative
generalized policy iteration (GPI) interpretation, we find that in practice a
\emph{single round of geometry-aware pair selection}, followed by multi-epoch
DPO optimization, yields more stable and effective personalization than repeated
re-selection. Accordingly, we present \textsc{GAP-DPO} primarily as a
three-step procedure:
(1) candidate exploration,
(2) utility-aware pair selection, and
(3) DPO optimization.
The overall procedure is summarized in Algorithm~\ref{alg:gap-dpo}.

\begin{algorithm}[t]
\caption{GAP-DPO: Off-Policy DPO with Utility-Aware Pair Selection}
\label{alg:gap-dpo}
\begin{algorithmic}[1]
\REQUIRE Dataset $\{(x_i, z_i)\}_{i=1}^N$; behavior policy $\pi_0$;
utility function $q(\cdot)$; DPO scale $\beta$; training epochs $K$.

\STATE \textbf{(A) Candidate generation (off-policy).}
\FOR{each prompt $(x_i, z_i)$}
    \STATE Sample a candidate pool $Y_i = \{y_{i,1}, \dots, y_{i,n}\}$ with
    $y_{i,j} \sim \pi_0(\cdot \mid x_i, z_i)$.
    \STATE Evaluate utilities $q(y_{i,j})$.
\ENDFOR

\STATE \textbf{(B) Utility-aware pair selection (one-shot).}
\FOR{each prompt $(x_i, z_i)$}
    \STATE Compute proxy DPO logit margins $\hat z_{ab}$ for candidate pairs $(y_a, y_b)$.
    \STATE Compute utility gaps $\Delta q_{ab} = q(y_a) - q(y_b)$.
    \STATE Select
    \[
    (y_i^+, y_i^-)
    \;=\;
    \arg\max_{(a,b)}
    \; \delta f_{ab},
    \quad
    \delta f_{ab}
    \;\approx\;
    \sigma(-\hat z_{ab}) \cdot \Delta q_{ab}.
    \]
    \STATE Add $(x_i, z_i, y_i^+, y_i^-)$ to dataset $\mathcal{D}$.
\ENDFOR

\STATE \textbf{(C) DPO optimization.}
Train $\pi_\theta$ for $K$ epochs by minimizing
\[
\mathcal{L}_{\mathrm{DPO}}(\theta;\mathcal{D}, \pi_{\mathrm{ref}}=\pi_0).
\]

\RETURN Trained policy $\pi_\theta$.
\end{algorithmic}
\end{algorithm}
\vspace{-5mm}

\paragraph{Main Result (Informal).}
\emph{Under mild geometric assumptions—specifically, bounded correlation between
score gradients and bounded policy drift—a geometry-aligned preference pair
selection induces a DPO update direction that yields a first-order improvement
in expected personalized utility.}

In particular, when the selected preference pair satisfies a gated utility gap
condition, the resulting DPO update lies within the ascent half-space of the
true utility gradient, leading to a local, trust-region-controlled improvement
in user utility. A formal statement and proof are provided in
Appendix~\ref{app:monotonicity}.
\vspace{-3mm}
\paragraph{Iterative Variants.}
The procedure above can be naturally extended to an iterative form, in which the
updated policy is periodically used to regenerate candidates and re-select
pairs, yielding a generalized policy iteration (GPI) loop. While this variant is
theoretically appealing, our empirical results indicate that repeated
re-selection provides limited additional benefit and may introduce instability
in personalized settings. In contrast, a single round of geometry-aware pair
selection followed by multi-epoch DPO optimization achieves comparable or better
performance with significantly improved stability. We therefore adopt the
one-shot selection variant as our default in experiments.

\vspace{-3mm}
\section{Experiments}
\subsection{Experiment Setting}

\paragraph{Datasets} To ensure data diversity, we choose various datasets from LongLaMP ~\cite{kumar2024longlamp}, LaMP ~\cite{salemi2023lamp}, and Amazon reviews~\cite{hou2024bridging, qiu2025measuring} for our experiments. In preliminary study, we use Amazon Review Datasets to evaluate different reinforcement learning algorithms and influence between synthetic negative samples and real user samples (See ~\ref{app:DPO-variants}~\ref{app:synthetic-vs-real}).
For the main comparison between our methods and baselines, we select Abstract Generation, Product Review, and Topic Writing from LongLaMP, as well as News Headline generation and Scholarly Title generation from LaMP.

In preparing the data, we follow the general setup of prior frameworks such as OPPU~\cite {tan2024democratizing} and CoPE~\cite{bu2025personalized} and select top 50 users with sufficient interaction histories as our evaluation cohort. For each user, we aggregate all historical interactions and then partition them into training, validation, and test subsets with an 8:1:1 ratio based on temporal ordering. When a personalization prompt requires leveraging user history as examples, we restrict retrieval to the training portion only, ensuring a realistic personalization setup without test leakage. In this case, we employ Contriever~\cite{izacard2021unsupervised} to retrieve the top-K most relevant history entries, which are then included as few-shot demonstrations for generating the target content. This design allows us to assess both the model’s raw personalization ability and its robustness to retrieved history length and quality. Dataset statistics and splits are provided in Table~\ref{tab:dataset_stats}

\begin{table}[t]
\centering
\small
\caption{Dataset statistics for the four evaluation tasks. AG = Abstract Generation, PR = Product Review, NH = News Headline, AB = Amazon Books.}
\label{tab:dataset_stats}
\begin{tabular}{lcccc}
\toprule
Dataset & AG & PR & NH & AB \\
\midrule
\# Questions        & 3,199 & 1,904 & 3,555 & 1,698 \\
Avg. Q Length      & 256.93 & 685.08 & 146.46 & 441.02 \\
Avg. Target Length & 152.08 & 343.25 & 10.07  & 283.46 \\
\bottomrule
\end{tabular}
\vspace{-3mm}
\end{table}

\vspace{-3mm}
\paragraph{Baseline Methods}
We compare our approach against two categories of baselines: SFT-only personalization methods and preference-based methods.
For SFT-only methods, we evaluate TAM~\cite{tan2024democratizing}, which targets to obtain a task-specific model by directly fine-tuning the model on user responses, and OPPU~\cite{tan2024democratizing}, which extends TAM by maintaining a separate LoRA adapter per user and continually training each adapter on the corresponding user's data.
For preference-based methods, we evaluate DPO~\cite{rafailov2023direct}, DICE~\cite{chen2024bootstrapping}, and CoPE~\cite{bu2025personalized}, each of which adopts different methods to construct preference data for training.
To construct preference pairs for DPO, we sample a negative example from the base model for each positive (ground-truth) instance.
For DICE, we adopt a simplified variant that removes the length penalty, using only its sampling-based preference construction method.
For CoPE, we similarly employ only its sampling strategy without per-user training, as the computational cost of per-user optimization is prohibitive at scale.
All preference-based methods are trained using the same preference data construction protocol to ensure fair comparison.

\vspace{-3mm}
\paragraph{Implementation Detail}
\textcolor{blue}{We compare our UG (Eq.~\ref{eq:delta_f_init}) and EG-UG (Eq.~\ref{eq:delta_f_proxy_simplified}) methods against the baselines above. We set the candidate pool size to 10 and use the ROUGE score against the gold response as the utility function for each training instance.} To ensure fairness,
all methods are implemented on LLaMA2-7B~\citep{touvron2023llama2} with a consistent LoRA configuration (rank = 8) and trained using the AdamW optimizer.
We first obtain the task-adapted model TAM by performing supervised fine-tuning with LoRA for 5 epochs at learning rate 1e-4, which serves as the LoRA baseline and the initialization for all preference-based methods.
OPPU continues training from TAM with per-user adapters.
For preference-based methods (DPO, DICE, CoPE, and our Methods), we continue training from TAM for an additional 5 epochs using preference optimization.
For RL algorithms, we initially experimented with different DPO variants and found that vanilla DPO performs competitively strong(see Appendix~\ref{app:DPO-variants} for detailed results); therefore, all preference-based methods are further trained for 5 epochs with standard DPO objective ($\beta=1.0$) and a learning rate of 5e-5.
To obtain diverse samples for preference pair construction, we set the generation temperature to 1.0.
All experiments are conducted on a single cluster node equipped with 4 NVIDIA H100 GPUs with 94 GB memory.
\providecommand{\codeavailability}{}
\codeavailability

\vspace{-3mm}
\paragraph{Evaluation Metrics}
In accordance with established practices in prior work~\cite{tan2024democratizing, kumar2024longlamp}, we employ a standard set of automatic metrics ROUGE-1, ROUGE-L~\cite{lin2004rouge}, and METEOR~\cite{banerjee2005meteor} to quantitatively assess the lexical overlap, fluency, and semantic correspondence between generated outputs and reference responses. 

\vspace{-3mm}
\subsection{Main Results}
\begin{table*}[t]
\centering
\caption{Performance comparison with different methods across four personalized content generation datasets.
All results are evaluated using ROUGE-1 (R-1), ROUGE-L (R-L), and METEOR (M).
\textbf{Best} results are in bold; \underline{underlined} values indicate the best baseline or second-best result.}
\label{tab:main_results}
\footnotesize
\setlength{\tabcolsep}{6pt}
\renewcommand{\arraystretch}{1.15}
\resizebox{\linewidth}{!}{%
\begin{tabular}{lcccccccccccc}
\toprule
& \multicolumn{3}{c}{Abstract Generation} 
& \multicolumn{3}{c}{Product Review} 
& \multicolumn{3}{c}{News Headline} 
& \multicolumn{3}{c}{Amazon Books} \\
\midrule
Method 
& R-1 & R-L & M 
& R-1 & R-L & M 
& R-1 & R-L & M 
& R-1 & R-L & M \\
\midrule
TAM 
& 0.3667 & 0.1900 & 0.2355 
& 0.4407 & 0.2754 & 0.3222 
& 0.2236 & 0.2048 & 0.1852 
& 0.3079 & 0.1378 & 0.2282 \\

OPPU
& 0.3753 & 0.1963 & 0.2368
    & 0.4597 & \underline{0.2923} & 0.3335
& \underline{0.2268} & \underline{0.2084} & \underline{0.1907}
& 0.3386 & \underline{0.1561} & 0.2537 \\

DPO 
& 0.2492 & 0.1096 & 0.2025 
& 0.3269 & 0.1269 & 0.2344 
& 0.1641 & 0.1519 & 0.1382 
& 0.3320 & 0.1353 & \textbf{0.3137} \\

CoPE 
& 0.2359 & 0.1072 & 0.1954 
& 0.3921 & 0.2258 & 0.3047 
& 0.1116 & 0.1009 & 0.1130 
& \underline{0.3412} & 0.1405 & \underline{0.3106} \\

DICE 
& \underline{0.4173} & \underline{0.2446} & \underline{0.2956} 
& \underline{0.4637} & 0.2917 & \underline{0.3572} 
& 0.1957 & 0.1810 & 0.1821 
& 0.3394 & 0.1460 & 0.2419 \\

UG (Ours)
& 0.4478 & 0.2526 & 0.3050
& 0.4964 & 0.3073 & 0.3701
& 0.2512 & 0.2298 & 0.2076
& 0.3829 & \textbf{0.1769} & 0.3011 \\

EG-UG (Ours)
& \textbf{0.4482} & \textbf{0.2548} & \textbf{0.3089}
& \textbf{0.4987} & \textbf{0.3117} & \textbf{0.3753}
& \textbf{0.2644} & \textbf{0.2385} & \textbf{0.2255}
& \textbf{0.3856} & 0.1756 & 0.2987 \\
\bottomrule
\end{tabular}%
}
\vspace{-3mm}
\end{table*}

First, we attempt to apply preference optimization methods to personalized content generation tasks.
However, preference optimization typically requires high-quality preference pairs, which are expensive and difficult to obtain in real-world personalization scenarios.
To enable scalable experimentation, we construct preference pairs by treating the original user's gold response as the preferred output and randomly sampling a response from the base model as the rejected output.
As shown in Table~\ref{tab:main_results}, this weakly-supervised preference signal leads to suboptimal performance: DPO underperforms SFT-based baselines such as TAM and OPPU across most tasks and metrics.
This observation reveals a key limitation of standard preference optimization under realistic data constraints and motivates our development of more robust user-guided optimization methods.

Table~\ref{tab:main_results} demonstrates that our methods consistently outperforms almost all baseline methods across four personalized content generation tasks. On average, the UG family improves ROUGE-1 by 13.1\% and ROUGE-L by 14.6\% relative to the best-performing baselines, with consistent gains also observed in METEOR scores, confirming the effectiveness and robustness of UG as a general personalization framework. Specifically, compared to SFT baselines (TAM and OPPU), our best-performing variant EG-UG achieves relative ROUGE-1 improvements ranging from 5.1\% to 18.2\%, with more pronounced gains on structurally complex tasks such as Abstract Generation and News Headline generation. Against stronger preference-based baselines (DPO, CoPE, and DICE), EG-UG delivers substantial improvements with ROUGE-1 margins of 7.4\% to 35.1\%, peaking at 35.1\% on the News Headline task under challenging low-context generation settings. Even the base UG variant consistently outperforms all baselines, achieving ROUGE-1 improvements between 7.2\% and 28.4\%. 

In summary, our methods outperforms across nearly all benchmarks,
achieving superior results across various datasets,
demonstrating the superior effectiveness of our approach.

\begin{figure}[t]
    \centering
    \begin{subfigure}[t]{0.48\linewidth}
        \centering
        \includegraphics[width=\linewidth]{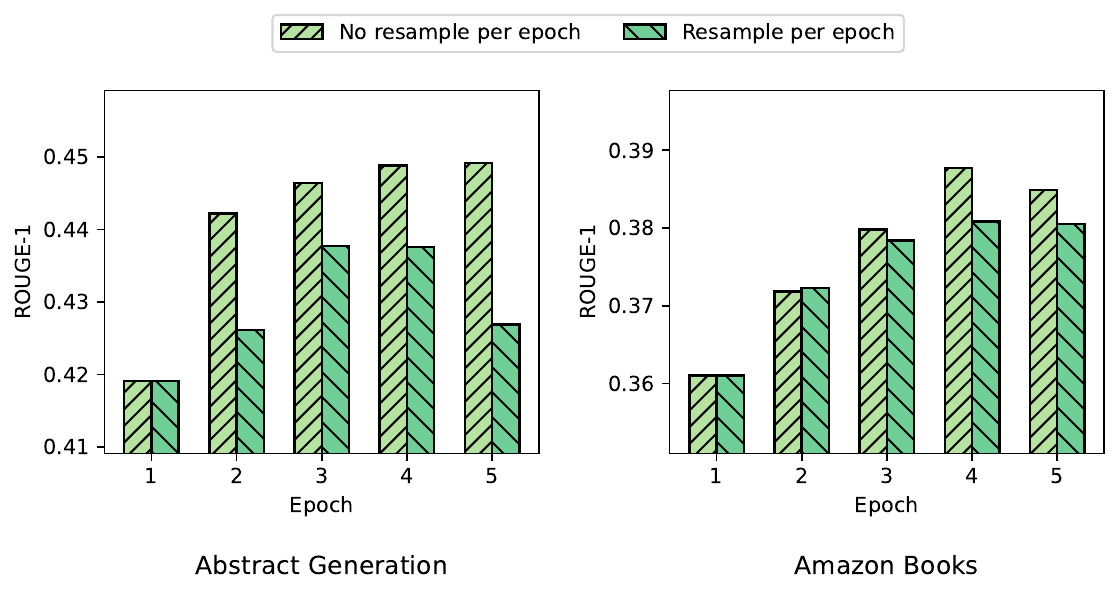}
        \caption{Comparison of non-resample and resample-per-epoch training strategies.}
        \label{fig:resample}
    \end{subfigure}
    \hfill
    \begin{subfigure}[t]{0.48\linewidth}
        \centering
        \includegraphics[width=\linewidth]{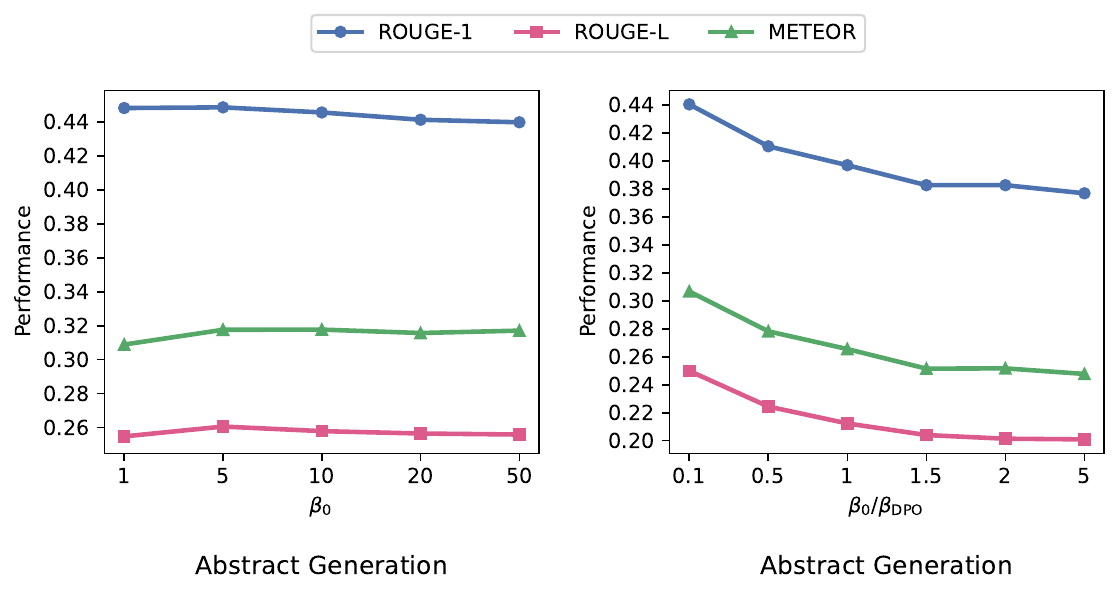}
        \caption{Sensitivity analysis of $\beta_0$ and $\beta_{\mathrm{DPO}}$. The left panel fixes $\beta_{\mathrm{DPO}}$ and varies $\beta_0$, while the right panel varies both parameters through their ratio $\beta_0 / \beta_{\mathrm{DPO}}$.}
        \label{fig:beta}
    \end{subfigure}
    \vspace{-2mm}
    \caption{Ablation studies of training strategies and hyperparameter sensitivity.}
    \label{fig:ablation_side_by_side}
    \vspace{-5mm}
\end{figure}

\paragraph{Resampling Strategies for Preference Construction} We first study whether our EG-UG method requires resampling of positive/negative candidates during training. We compare two training strategies: non-resample per epoch, where a fixed set of positive–negative preference pairs is used throughout training, and resample per epoch, where candidate training pairs are regenerated and re-scored after each epoch. As shown in Fig~\ref{fig:resample}, as the number of training epochs increases, the performance of both strategies consistently improves on Amazon Books dataset. However, on Abstract Generation, the no-resample strategy reaches near-optimal performance by the second epoch, with only marginal improvements from additional training. Across both datasets, the resample-per-epoch strategy does not show clear advantages over using a fixed preference set. Overall, these results indicate that EG-UG does not require per-epoch resampling, concluding that training for multiple epochs on a fixed set of preference pairs is sufficient to achieve strong performance while avoiding the additional computational overhead of resampling.

\vspace{-3mm}
\paragraph{Effect on $\beta$} In Eq.~\ref{eq:dpo_direction}, the $\beta$ value for calculating $z_{\theta}$ is identical to the $\beta$ used in DPO. In practice, however, these two $\beta$ values can be decoupled to provide additional flexibility. We therefore study the impact and relation between the two hyperparameters, where we temporarily denote them as $\beta_{\text{DPO}}$ and $\beta_0$, respectively.
In Figure~\ref{fig:beta}, we report the results on two complementary settings: In the left panel, we fix $\beta_{\text{DPO}}$ and vary $\beta_0$. We observe that the performance remains largely stable across a wide range of $\beta_0$ values, with only minor variations in ROUGE-1, ROUGE-L, and METEOR. This suggests that, when the $\beta_{DPO}$ is fixed, the method is relatively insensitive to the exact choice of $\beta_0$. In the right panel, we jointly vary the two parameters through their ratio $\beta_0 / \beta_{\text{DPO}}$. As this ratio increases, performance consistently degrades across all metrics, indicating that large mismatches between $\beta_0$ and $\beta_{\text{DPO}}$ can negatively affect preference learning. Actually, these analysis guided us to simply choose $\beta_{0}=\beta_{DPO}$ in all our experiments.

\vspace{-3mm}
\section{Conclusion}

In this work, we studied personalized preference optimization through the lens of optimization geometry. We showed that the effectiveness of Direct Preference Optimization (DPO) in personalized settings critically depends not only on the objective itself, but on how preference pairs are selected. By analyzing the first-order interaction between DPO updates and the gradient of expected user utility, we identified gradient alignment as a key mechanism governing whether preference optimization advances or hinders personalization.

Motivated by this analysis, we proposed \textsc{GAP-DPO}, a geometry-aware framework that performs utility-aligned pair selection under off-policy training. Our theoretical results establish that, under mild geometric and trust-region assumptions, geometry-aligned pair selection ensures that DPO updates induce local improvement in expected personalized utility. Empirically, we demonstrated that \textsc{GAP-DPO} consistently improves stylistic fidelity, preference alignment, and generation quality across a range of personalized text generation benchmarks.

Beyond the specific algorithm, our findings suggest a broader perspective on preference learning: pair selection is not a heuristic preprocessing step, but an intrinsic component of the optimization geometry. Viewing preference optimization as a directional alignment problem provides a unifying framework for understanding stability, personalization divergence, and the role of confidence in contrastive learning.

Future work may extend this framework to richer preference structures, such as multi-way comparisons, hierarchical or graph-based preference representations, and interactive or online personalization. We also see opportunities to integrate geometry-aware selection with safety constraints, diversity objectives, and human-in-the-loop feedback. 

\newpage
\section*{Impact Statements}
This work studies personalized preference optimization for large language models, with a focus on improving alignment between model behavior and user-specific utilities. By introducing geometry-aware pair selection, our approach aims to enhance personalization quality, stylistic fidelity, and user satisfaction while maintaining training stability and scalability.

\paragraph{Potential Benefits.}
The proposed framework may enable more effective and efficient personalization of language models across diverse users and applications. Improved alignment with individual preferences could enhance user experience in domains such as creative writing, education, accessibility tools, and assistive technologies. By formalizing preference optimization as a geometry-aligned process, this work also contributes theoretical insights that may benefit future research on stable and interpretable alignment methods.

\paragraph{Potential Risks.}
Personalized language models may amplify existing biases or reinforce narrow user preferences if not carefully designed. Over-personalization could lead to echo chambers or reduce exposure to diverse viewpoints. Additionally, reliance on external utility functions or automated evaluators may introduce unintended biases or misalignment if such utilities are imperfect or poorly calibrated.

\paragraph{Mitigations.}
Our method does not prescribe a specific utility function and can incorporate safeguards such as regularization, diversity constraints, or human-in-the-loop evaluation to mitigate overfitting or bias amplification. The geometry-aware pair selection framework is compatible with existing safety mechanisms, including trust-region constraints and reference policies, which help prevent uncontrolled model drift. We encourage practitioners to combine personalization with transparency and oversight when deploying such systems.

Overall, this work advances understanding of preference optimization while highlighting the importance of responsible personalization and careful utility design.

\bibliographystyle{plainnat}
\bibliography{ref,www2026}


\appendix
\newpage
\appendix
\onecolumn

\section{GAP-DPO Analysis}
\label{app:monotonicity}

This appendix provides a local, geometry-driven justification for why \textbf{GAP-DPO}
tends to increase expected utility under mild regularity assumptions.
Fix a prompt--user context $(x,z)$ and define the conditional expected utility
\[
f(\theta)\;=\;\mathbb{E}_{y\sim \pi_\theta(\cdot\mid x,z)}[q(y)] .
\]
GAP-DPO selects a preference pair $(y^+,y^-)$ from an off-policy candidate set and then
applies a DPO update. The key idea is to select pairs such that the resulting DPO update
direction lies in (or close to) the ascent half-space of a \emph{utility-gradient proxy}.
A central point in the analysis is to \emph{separate}:

\begin{enumerate}
\item \textbf{i.i.d.\ off-policy sampling (unbiased estimation):} if $(y^+,y^-)$ are sampled
i.i.d.\ from a reference policy $\pi_{\mathrm{ref}}$, then a simple importance-weighted
Monte Carlo proxy is unbiased for $\nabla_\theta f(\theta)$.

\item \textbf{pool-based pair selection (biased but meaningful):} if $(y^+,y^-)$ are chosen
by a selection rule $S$ from a candidate pool, then the same proxy is generally biased for
$\nabla_\theta f(\theta)$. This is not a flaw; it means the algorithm is performing local ascent on a
\emph{selection-tilted objective} induced by $S$ on the sampled support.
\end{enumerate}

Finally, in a high-dimensional regime where gradients associated with distinct prompt-pairs
are approximately orthogonal, \emph{per-pair} positive alignment implies \emph{batch} positive alignment,
yielding a clean route to monotonic first-order improvement.

\subsection{Lemma 1: Proxy Gradient Alignment}
\label{lem:alignment}

Let the score function be
\[
g(y)\;=\;\nabla_\theta \log \pi_\theta(y\mid x,z),
\]
and let $g^+,g^-$ correspond to the two responses $y^+,y^-$ (for the same $(x,z)$).
Define importance weights w.r.t.\ an off-policy reference $\pi_{\mathrm{ref}}$:
\[
w(y)\;=\;\frac{\pi_\theta(y\mid x,z)}{\pi_{\mathrm{ref}}(y\mid x,z)},
\qquad
w^\pm := w(y^\pm),
\quad
q^\pm := q(y^\pm).
\]
Let $d_{\mathrm{DPO}}=-\nabla_\theta \mathcal{L}_{\mathrm{DPO}}(\theta)$ denote the (single-pair) DPO update direction,
\[
d_{\mathrm{DPO}}
=
\beta\,\sigma(-z_\theta)\,(g^+-g^-),
\qquad
z_\theta
=
\log \pi_\theta(y^+\mid x,z)-\log \pi_\theta(y^-\mid x,z).
\]

\begin{lemma}[Exact Proxy Alignment Identity]
\label{lem:proxy-alignment}
Define the two-sample utility-gradient proxy
\[
\widehat{\nabla_\theta f}
\;\triangleq\;
\frac12\Big(w^+ q^+ g^+ + w^- q^- g^-\Big),
\]
and its alignment with the DPO update direction
\[
\widehat{\delta f}
\;\triangleq\;
\big\langle \widehat{\nabla_\theta f},\, d_{\mathrm{DPO}} \big\rangle .
\]
Then $\widehat{\delta f}$ satisfies the exact identity
\begin{align}
\widehat{\delta f}
&=
\frac{\beta}{2}\,\sigma(-z_\theta)\Big(
w^+q^+\|g^+\|^2
-
w^-q^-\|g^-\|^2
-
(w^+q^+-w^-q^-)\langle g^+,g^-\rangle
\Big).
\label{eq:proxy-alignment-identity}
\end{align}
\end{lemma}

\paragraph{Proof.}
By definition,
\[
\widehat{\delta f}
=
\left\langle
\frac12(w^+q^+g^+ + w^-q^-g^-),\;
\beta\sigma(-z_\theta)(g^+-g^-)
\right\rangle .
\]
Expanding by bilinearity and collecting norm-squared and cross terms yields
\eqref{eq:proxy-alignment-identity}.
\hfill$\square$

\paragraph{Interpretation.}
If $\langle g^+,g^-\rangle$ is small (e.g., locally orthogonal scores), then the sign of
$\widehat{\delta f}$ is dominated by the weighted norm gap
$w^+q^+\|g^+\|^2-w^-q^-\|g^-\|^2$. This is precisely what GAP-DPO exploits by selecting
pairs that yield large positive $\widehat{\delta f}$.

\subsection{Unbiased Utility-Gradient Estimation vs.\ Pool-Based Pair Selection}
\label{app:unbiased-vs-selection}

The expected utility satisfies the score-function identity
\[
\nabla_\theta f(\theta)
=
\mathbb{E}_{y\sim\pi_\theta(\cdot\mid x,z)}\!\left[q(y)\,g(y)\right].
\]
Under off-policy sampling from $\pi_{\mathrm{ref}}$, the standard importance-weighted form is
\[
\nabla_\theta f(\theta)
=
\mathbb{E}_{y\sim\pi_{\mathrm{ref}}(\cdot\mid x,z)}\!\left[w(y)\,q(y)\,g(y)\right].
\]

\paragraph{Regime A: i.i.d.\ off-policy sampling (unbiased Monte Carlo).}
Assume $y^+,y^- \overset{\text{i.i.d.}}{\sim} \pi_{\mathrm{ref}}(\cdot\mid x,z)$ (with replacement) and define
\[
\widehat{\nabla_\theta f}
=
\frac12\Big(w^+q^+g^+ + w^-q^-g^-\Big).
\]
Then $\widehat{\nabla_\theta f}$ is unbiased:
\[
\mathbb{E}\!\left[\widehat{\nabla_\theta f}\right]
=
\mathbb{E}_{y\sim\pi_{\mathrm{ref}}}\!\left[w(y)\,q(y)\,g(y)\right]
=
\nabla_\theta f(\theta).
\]
In this regime, Lemma~\ref{lem:proxy-alignment} gives an explicit decomposition of
$\widehat{\delta f}=\langle \widehat{\nabla_\theta f}, d_{\mathrm{DPO}}\rangle$.
If one wishes to interpret $\mathbb{E}[\widehat{\delta f}]$ as a proxy for expected improvement,
it is cleanest to evaluate $\widehat{\nabla_\theta f}$ on an \emph{independent replica} of samples
to avoid coupling between the update direction and the estimator.

\paragraph{Regime B: pool-based pair selection (biased, but a local ascent proxy).}
In GAP-DPO, one generates a candidate pool of size $n$,
\[
\mathcal{Y}=\{y_1,\dots,y_n\},
\qquad y_j\sim \pi_{\mathrm{ref}}(\cdot\mid x,z),
\]
then applies a (possibly deterministic) selection rule $S$ to pick a pair
\[
(y^+,y^-)=S(\mathcal{Y}).
\]
Then $(y^+,y^-)$ are not i.i.d.\ from $\pi_{\mathrm{ref}}$; their law is the selection-induced distribution
\[
(y^+,y^-)\sim P_S(\cdot\mid x,z),
\]
which depends on $\pi_{\mathrm{ref}}$ and $S$. Consequently,
\[
\mathbb{E}_{(y^+,y^-)\sim P_S}\!\left[\widehat{\nabla_\theta f}\right]\neq \nabla_\theta f(\theta)
\quad \text{in general.}
\]
This bias is inherent: selection concentrates updates on an ``informative'' region of the sampled support.
Accordingly, $\widehat{\nabla_\theta f}$ should be interpreted as a \emph{local ascent proxy on the selected support},
or equivalently as a gradient estimator for a \emph{tilted objective} that treats $P_S$ as locally frozen.

\paragraph{Why Lemma~\ref{lem:proxy-alignment} still matters under selection.}
Lemma~\ref{lem:proxy-alignment} is algebraic and holds for any realized pair, regardless of how it was obtained.
Thus, even under selection, \eqref{eq:proxy-alignment-identity} provides a computable directional certificate:
it explicitly connects the DPO direction $d_{\mathrm{DPO}}$ to a utility-weighted score proxy in the same basis.
This is exactly what enables GAP-DPO to enforce per-pair ascent conditions such as $\widehat{\delta f}>0$.

\subsection{From Positive Alignment to Local Improvement (First-Order)}
\label{app:firstorder-improvement}

We record a standard smoothness implication that translates positive alignment into a
local improvement guarantee.

\begin{assumption}[$L$-smooth objective]
\label{ass:smooth}
The objective $f(\theta)$ is differentiable and $L$-smooth:
$\|\nabla f(\theta')-\nabla f(\theta)\|\le L\|\theta'-\theta\|$ for all $\theta,\theta'$.
\end{assumption}

\begin{lemma}[Local improvement under positive alignment]
\label{lem:local-improve}
Under Assumption~\ref{ass:smooth}, for any direction $d$ and step size $\eta>0$,
\[
f(\theta+\eta d)\;\ge\; f(\theta)+\eta\langle \nabla f(\theta), d\rangle -\frac{L\eta^2}{2}\|d\|^2 .
\]
In particular, if $\langle \nabla f(\theta), d\rangle \ge \gamma>0$, then choosing
$\eta \le \gamma/(L\|d\|^2)$ yields $f(\theta+\eta d)\ge f(\theta)$.
\end{lemma}

Although $q(y)$ may be non-differentiable with respect to the discrete output $y$, the objective
\begin{equation}
f(\theta) = \mathbb{E}_{y \sim \pi_\theta}\big[q(y)\big]
\end{equation}
is differentiable in $\theta$ since the policy $\pi_\theta$ is parameterized by smooth softmax mappings. Under the additional assumptions of bounded utility and a KL trust-region constraint that restricts policy drift, $f$ is locally smooth in a neighborhood of $\theta$ and therefore admits a second-order Taylor expansion with bounded remainder.

\paragraph{Remark (proxy-based use).}
In GAP-DPO we do not have $\nabla f(\theta)$. Instead, we select pairs so that
$\widehat{\delta f}=\langle \widehat{\nabla f}, d_{\mathrm{DPO}}\rangle$ is positive and large,
making $\langle \nabla f(\theta), d_{\mathrm{DPO}}\rangle$ more likely to be positive in expectation
(Regime A) or for the tilted objective induced by selection (Regime B).

\subsection{Informal Takeaway}
\label{app:informal-takeaway}

The above results are deliberately \emph{local} and \emph{geometry-driven}.
They do not claim every selected pair improves the global objective.
Rather, they explain why \textbf{pair selection is a principled mechanism for producing updates that tend to increase utility}:

\begin{enumerate}
\item A small update improves $f$ to first order when $\langle \nabla f(\theta), d\rangle>0$
(Lemma~\ref{lem:local-improve}).

\item Lemma~\ref{lem:proxy-alignment} provides an exact, computable decomposition of the proxy alignment
$\widehat{\delta f}=\langle \widehat{\nabla f}, d_{\mathrm{DPO}}\rangle$ for any realized pair.

\item In the i.i.d.\ regime, $\widehat{\nabla f}$ is unbiased for $\nabla f$, so filtering for $\widehat{\delta f}>0$
acts as a sign-filtering / variance-reduction heuristic that increases the expected first-order gain.

\item In the pool-selection regime, the proxy becomes biased because the algorithm optimizes a selection-tilted
objective on the sampled support; enforcing $\widehat{\delta f}>0$ is still consistent with local ascent for that
tilted objective.

\end{enumerate}

GAP-DPO does not require access to $\nabla f$. It selects pairs that maximize a computable alignment proxy
$\widehat{\delta f}$ expressed in the same score-function basis as the DPO gradient. Positive proxy alignment
increases the likelihood of a positive first-order utility gain.

\section{Additional Experiments}
\subsection{Comparison of DPO Variants}
\label{app:DPO-variants}

\begin{table}[t]
\centering
\caption{Preliminary comparison of reinforcement learning algorithms on two Amazon review generation tasks.}
\label{tab:dpo_variants}
\scriptsize
\setlength{\tabcolsep}{5pt}
\renewcommand{\arraystretch}{1.1}
\begin{tabular}{lcccccc}
\toprule
& \multicolumn{3}{@{}c@{}}{Movies \& TV} 
& \multicolumn{3}{@{}c@{}}{CDs \& Vinyl} \\
\cmidrule(lr){2-4} \cmidrule(lr){5-7}
Method 
& R-1 & R-L & M 
& R-1 & R-L & M \\
\midrule
DPO  
& \textbf{0.3551} & 0.1540 & \textbf{0.2748} 
& \textbf{0.3712} & 0.1653 & \textbf{0.2748} \\

IPO  
& 0.2521 & 0.1173 & 0.1680 
& 0.1606 & 0.0887 & 0.1104 \\

CPO  
& 0.3195 & 0.1371 & 0.2516 
& 0.3588 & 0.1586 & 0.2673 \\

KTO  
& 0.3523 & \textbf{0.1564} & 0.2567 
& 0.3637 & \textbf{0.1660} & 0.2557 \\

ORPO 
& 0.3448 & 0.1530 & 0.2449 
& 0.3537 & 0.1635 & 0.2410 \\
\bottomrule
\end{tabular}
\end{table}

We conduct preliminary experiments comparing standard DPO with several recently proposed DPO variants, including IPO~\cite{azar2024general}, CPO~\cite{achiam2017constrained}, KTO~\cite{ethayarajh2024kto}, and ORPO~\cite{hong2024orpo}, on two Amazon review generation tasks.
As shown in Table~\ref{tab:dpo_variants}, we observe that these variants do not consistently outperform standard DPO across evaluation metrics.
In several cases, they yield noticeably worse performance, particularly in terms of ROUGE-L and METEOR.

Given the lack of consistent improvements and for experimental clarity, we adopt standard DPO as the reinforcement learning algorithm in all subsequent experiments.

\subsection{Synthetic Negatives vs. Real Negatives}\label{app:synthetic-vs-real}
\begin{table*}[t]
\centering
\caption{Performance comparison between synthetic Negatives vs. real-user Negatives on Amazon Book Dataset.
We compare using other users' real reviews as negatives versus LLM-generated synthetic negatives.}
\label{tab:amazon_books_negatives}
\scriptsize
\setlength{\tabcolsep}{5pt}
\renewcommand{\arraystretch}{1.1}
\begin{tabular}{lcccccc}
\toprule
& \multicolumn{3}{@{}c@{}}{Other Users' Reviews as Negatives} 
& \multicolumn{3}{@{}c@{}}{LLM-Generated Negatives} \\
\cmidrule(lr){2-4} \cmidrule(lr){5-7}
Method 
& R-1 & R-L & M 
& R-1 & R-L & M \\
\midrule
TAM 
& 0.3079 & 0.1378 & 0.2282 
& 0.3079 & 0.1378 & 0.2282 \\

OPPU 
& 0.3654 & 0.1784 & 0.2902 
& 0.3654 & 0.1784 & 0.2902 \\

DPO 
& 0.3597 & 0.1520 & 0.3233 
& 0.3320 & 0.1353 & 0.3137 \\

CoPE 
& 0.3333 & 0.1519 & 0.2731 
& 0.3412 & 0.1405 & 0.3106 \\

DICE 
& 0.3023 & 0.1223 & 0.2448 
& 0.3394 & 0.1460 & 0.2419 \\

UG (Ours) 
& 0.3770 & 0.1655 & 0.3394 
& 0.3829 & 0.1769 & 0.3011 \\

EG-UG (Ours) 
& 0.3748 & 0.1618 & 0.3438 
& 0.3856 & 0.1756 & 0.2987 \\
\bottomrule
\end{tabular}
\end{table*}

Table~\ref{tab:amazon_books_negatives} investigates the impact of different negative sampling sources on the Amazon Books dataset.
In many real-world personalization scenarios, collecting high-quality negative examples from other users' real reviews is often infeasible or prohibitively expensive.
As a result, a practical alternative is to construct synthetic negatives directly from LLM-generated outputs.
This experiment evaluates whether such synthetic data can effectively substitute real negatives in preference-based optimization.

Across most baseline methods, using LLM-generated negatives leads to comparable  performance compared to real-review negatives.
Specifically, our UG family demonstrates strong robustness to the choice of negative samples.
Notably, both UG and EG-UG achieve equal or even better performance when trained with synthetic negatives, with EG-UG attaining the best ROUGE-1 and competitive ROUGE-L and METEOR scores under LLM-generated negatives.

These results suggest that, under the proposed UG framework, synthetically generated negatives can serve as effective—and sometimes superior—substitutes for real user reviews.
This finding highlights the practicality of our approach, enabling scalable preference-based personalization even in settings where real negative data is unavailable.

\subsection{Results with OPPU framework}
\begin{table*}[t]
\centering
\scriptsize
\caption{Comparison of different sampling strategies under the OPPU framework across three content generation tasks.}
\label{tab:oppu2}

\begin{tabular}{lccccccccc}
\toprule
& \multicolumn{3}{c}{Abstract Generation} 
& \multicolumn{3}{c}{Product Review} 
& \multicolumn{3}{c}{News Headline} \\
\cmidrule(lr){2-4}
\cmidrule(lr){5-7}
\cmidrule(lr){8-10}

Method 
& R-1 & R-L & METEOR 
& R-1 & R-L & METEOR 
& R-1 & R-L & METEOR \\
\midrule
OPPU 
& 0.3753 & 0.1963 & 0.2368 
& 0.4597 & 0.2923 & 0.3335 
& 0.2268 & 0.2084 & 0.1907 \\

DICE 
& 0.3726 & 0.1969 & 0.2340 
& 0.4633 & 0.2908 & 0.3424 
& 0.2321 & 0.2078 & 0.1886 \\

CoPE 
& 0.3695 & 0.1909 & 0.2373 
& 0.4203 & 0.2337 & 0.3018 
& 0.1997 & 0.1854 & 0.1686 \\

UG (Ours) 
& 0.3898 & \textbf{0.2105} & 0.2570 
& 0.4817 & \textbf{0.3047} & \textbf{0.3580} 
& 0.2319 & 0.2121 & 0.1959 \\

EG-UG (Ours) 
& \textbf{0.3906} & 0.2071 & \textbf{0.2576} 
& \textbf{0.4821} & 0.3024 & 0.3575 
& \textbf{0.2416} & \textbf{0.2246} & \textbf{0.2018} \\

\bottomrule
\end{tabular}
\end{table*}

We further investigate the efficacy of our methods under the OPPU framework.
Table ~\ref{tab:oppu2} and compare different sampling strategies within the OPPU framework with different training epochs. Overall, our UG and EG-UG approaches consistently outperform prior sampling methods such as OPPU and CoPE, and are competitive with or better than DICE on most metrics. In particular, UG achieves the strongest performance on Product Review, while UG/EG attain the best results on Abstract Generation, suggesting that more user-aligned sampling can yield more effective personalization when plugged into the OPPU pipeline.

Compared with the main results in Table ~\ref{tab:main_results}, we find that our methods do not rely on the OPPU framework to perform well, and the additional gains from per-user reinforcement learning are often limited. A plausible explanation is that each user has only a small amount of personalized data, making preference signals noisy and hard to estimate reliably; under such data-scarce conditions, per-user reinforcement learning training may provide diminishing returns. In contrast, our methods emphasize constructing higher-quality training signals from limited data, which can be more impactful than increasing optimization complexity in practical personalization settings.

\section{Prompt Template and Case Study}
\subsection{Prompt Template}
We provide the prompt template we use in experiemtents as below:

\vspace{6pt}
\noindent
\colorbox[gray]{0.9}{
  \parbox{\dimexpr\linewidth-2\fboxsep\relax}{
  \textbf{Prompt Template For Abstract Generation}
  }
}
\vspace{-2pt}
\noindent
\setlength{\fboxsep}{6pt}
\setlength{\fboxrule}{0.4pt}
\doublebox{
\begin{minipage}{0.92\linewidth}
\small
You are an \textbf{academic researcher}. \\  
Your task is to generate an academic-style abstract that matches the author's writing style based on the paper's abstract.  \\
Here are reference examples (for style/tone ONLY; do not copy them): \texttt{<REFERENCE\_ABSTRACT>} is the abstract of \texttt{<REFERENCE\_TITLE>}.\\
Generate a NEW abstract for the paper titled: \texttt{<TARGET\_TITLE>}\\ 
\textbf{Constraints:}
\begin{itemize}
  \item DO NOT copy any sentences from the reference abstracts.
  \item Length:150-250 words; formal, concise, and self-contained.
  \item Suggested structure: background/objective → method → data/pipeline → results/impact → (optional) deployment/cloud aspects.
  \item Write only the abstract text, without headings or extra commentary.
\end{itemize}
\end{minipage}
}

\subsection{Quantitative Study: }
To quantitative illustrate the generation result of our approach over
baseline approaches, we present a few representative examples from
abstract generation, and leverage GPT-5 to evaluate the output of both baseline and our methods based on 3 aspects: Ground Truth Alignment, Structural Organization Alignment, and Stylistic \& Rhetorical Alignment, and finally give each generated out a score as an overall evaluation with below prompt:

\newpage
\vspace{6pt}
\noindent
\colorbox[gray]{0.9}{
  \parbox{\dimexpr\linewidth-2\fboxsep\relax}{
  \textbf{LLM as Judge Evaluation Prompt}
  }
}
\vspace{-2pt}
\noindent
\setlength{\fboxsep}{6pt}
\setlength{\fboxrule}{0.4pt}
\doublebox{
\begin{minipage}{0.92\linewidth}
\small
Role:
You are an expert academic reviewer evaluating a generated research abstract.

Task:
Evaluate the generated abstract along two independent dimensions.
Treat these dimensions as orthogonal and do not assume they correlate.

---

Dimension 1: Ground Truth Alignment (0–10)
Compare the generated abstract to the reference (ground-truth) abstract and assess alignment across:

Semantic Content Alignment
– Core ideas, claims, scope, and technical contributions
– Inclusion or omission of key concepts

Structural Organization Alignment
– High-level structure (e.g., problem → method → results → implications)
– Ordering and emphasis of ideas

Stylistic \& Rhetorical Alignment
– Tone consistency relative to the reference
– Lexical alignment (terminology, phrasing)
– Rhetorical framing (what is emphasized or de-emphasized)
– Persona coherence (authorial voice and positioning)

Scoring (0–10):

9–10: Near-paraphrase; same content, structure, and style
6–8: Same core ideas with moderate structural or stylistic drift
3–5: Partial overlap; missing or distorted key elements
0–2: Largely unrelated

---

Dimension 2: Intrinsic Abstract Quality (Reference-Agnostic) (0–10)
Evaluate the abstract on its own merits, ignoring the ground truth entirely:

– Clarity and readability
– Logical coherence and flow
– Conciseness and precision
– Stylistic consistency (stable tone, voice, and level of formality)
– Overall academic writing quality

Scoring (0–10):

9–10: Clear, fluent, stylistically consistent, publication-ready
6–8: Generally strong with minor issues
3–5: Understandable but weakly written or stylistically unstable
0–2: Poorly written or incoherent

---

Output Format
Provide:

Ground Truth Alignment score (0–10)
– 2–3 sentence justification
Intrinsic Abstract Quality score (0–10)
– 2–3 sentence justification

Clearly label both scores so they can be tabulated.

---

Important Notes
The two scores need not correlate.
A response may be well written but poorly aligned, or well aligned but stylistically weak.
Focus explicitly on style and persona consistency, not just semantic correctness.

\end{minipage}
}

\renewcommand{\arraystretch}{1.15}

\begin{longtable}{p{3.2cm}cc cc cc cc cc}
\caption{Alignment and intrinsic quality scores across abstract generation tasks.}
\label{tab:fancy_scores}\\
\toprule
\multicolumn{11}{l}{\textbf{Abstract Generation Tasks}} \\
\midrule
\textbf{Method} 
& \multicolumn{2}{c}{\textbf{Task 1}} 
& \multicolumn{2}{c}{\textbf{Task 2}} 
& \multicolumn{2}{c}{\textbf{Task 3}} 
& \multicolumn{2}{c}{\textbf{Task 4}} 
& \multicolumn{2}{c}{\textbf{Task 5}} \\
\cmidrule(lr){2-3}
\cmidrule(lr){4-5}
\cmidrule(lr){6-7}
\cmidrule(lr){8-9}
\cmidrule(lr){10-11}
& \textbf{A} & \textbf{Q}
& \textbf{A} & \textbf{Q}
& \textbf{A} & \textbf{Q}
& \textbf{A} & \textbf{Q}
& \textbf{A} & \textbf{Q} \\
\midrule
\endfirsthead

\toprule
\multicolumn{11}{l}{\textbf{Abstract Generation Tasks}} \\
\midrule
\textbf{Method} 
& \multicolumn{2}{c}{\textbf{Task 1}} 
& \multicolumn{2}{c}{\textbf{Task 2}} 
& \multicolumn{2}{c}{\textbf{Task 3}} 
& \multicolumn{2}{c}{\textbf{Task 4}} 
& \multicolumn{2}{c}{\textbf{Task 5}} \\
\cmidrule(lr){2-3}
\cmidrule(lr){4-5}
\cmidrule(lr){6-7}
\cmidrule(lr){8-9}
\cmidrule(lr){10-11}
& \textbf{A} & \textbf{Q}
& \textbf{A} & \textbf{Q}
& \textbf{A} & \textbf{Q}
& \textbf{A} & \textbf{Q}
& \textbf{A} & \textbf{Q} \\
\midrule
\endhead

\bottomrule
\endfoot

TAM
& \scorecell{8} & \scorecell{7}
& \scorecell{1} & \scorecell{6}
& \scorecell{8} & \scorecell{7}
& \scorecell{8} & \scorecell{8}
& \scorecell{8} & \scorecell{8} \\

OPPU
& \scorecell{6} & \scorecell{8}
& \scorecell{1} & \scorecell{5}
& \scorecell{7} & \scorecell{6}
& \scorecell{7} & \scorecell{7}
& \scorecell{9} & \scorecell{8} \\

DPO
& \scorecell{3} & \scorecell{4}
& \scorecell{5} & \scorecell{3}
& \scorecell{3} & \scorecell{2}
& \scorecell{3} & \scorecell{3}
& \scorecell{3} & \scorecell{2} \\

CoPE
& \scorecell{1} & \scorecell{5}
& \scorecell{2} & \scorecell{3}
& \scorecell{2} & \scorecell{3}
& \scorecell{4} & \scorecell{4}
& \scorecell{1} & \scorecell{4} \\

DICE
& \scorecell{4} & \scorecell{8}
& \scorecell{6} & \scorecell{6}
& \scorecell{5} & \scorecell{7}
& \scorecell{6} & \scorecell{7}
& \scorecell{5} & \scorecell{7} \\

UG (Ours)
& \scorecell{6} & \scorecell{7}
& \scorecell{9} & \scorecell{8}
& \scorecell{8} & \scorecell{8}
& \scorecell{9} & \scorecell{8}
& \scorecell{8} & \scorecell{8} \\

EG-UG (Ours)
& \scorecell{7} & \scorecell{8}
& \scorecell{8} & \scorecell{7}
& \scorecell{9} & \scorecell{8}
& \scorecell{9} & \scorecell{8}
& \scorecell{9} & \scorecell{8} \\

\end{longtable}

Table ~\ref{tab:fancy_scores} reports LLM-as-a-judge scores across different tasks. Overall, EG-UG achieves the strongest and most consistent performance among all methods. Compared with prior approaches such as TAM, OPPU, and DPO variants, EG-UG attains higher scores on the majority of tasks in both alignment and quality dimensions, and shows clear improvements over UG, indicating the effectiveness of incorporating explicit guidance into the UG framework. Notably, EG-UG ranks first or second on most tasks, demonstrating its ability to generate abstracts that are not only well-aligned with task requirements but also of high intrinsic quality. These results suggest that EG-UG provides a robust and balanced improvement across diverse abstract generation settings.

\newtcolorbox{abstractbox}[3]{%
  breakable,
  enhanced,
  colback=gray!3,
  colframe=black!40,
  boxrule=0.6pt,
  arc=2mm,
  left=2mm,right=2mm,top=1.5mm,bottom=1.5mm,
  title=\textbf{#1}\hfill{\small \textbf{Align:} #2\quad \textbf{Quality:} #3},
  fonttitle=\normalsize,
}

\subsubsection{Task 1 Result}

\begin{abstractbox}{Ground Truth (Reference)}{--}{--}
Detection and description of local image features has proven to be a powerful paradigm for a variety of applications in computer vision. Often, this process includes an orientation assignment step to render the overall process invariant to in-plane rotation. In this paper, we review several different existing algorithms and propose two novel, efficient methods for orientation assignment. The first method exhibits a very good speed-performance trade-off; the second is capable of multiple orientations and performs comparable to SIFT's orientation assignment while being significantly cheaper. Additionally, we improve one of the existing orientation assignment methods by generalizing it. All algorithms are evaluated empirically under a variety of conditions and in combination with six keypoint detectors.
\end{abstractbox}

\begin{abstractbox}{TAM}{8}{7}
Visual object recognition relies heavily on accurate detection and correspondence between visually similar objects observed at multiple viewpoints. A key technique used for viewpoint consistency is to project images of interest into a reference frame, called a canonical space. In this framework, high dimensional representations can be exploited to extract orientation constraints for each keypoint location and orientation. These constraints are then solved to yield orientations for all keypoint locations which are consistent with existing object identities. However, errors in the initial orientation assignment can cause failure in constructing proper constraint matrices. In this work, we present a method for refining orientation assignments in a fast and efficient way such that the error in orientation assignment affects the result less. The idea is to keep two copies of the solution at different iterations such that, at convergence, the goal of finding the global minimum has been achieved even when, initially, the wrong local minima is reached. We report the experimental results on VOT'97, OTB'06 and LCVG'10 datasets showing an improvement in both speed and accuracy.
\end{abstractbox}

\begin{abstractbox}{OPPU}{6}{8}
The orientation of the projection of keypoints is important to accurate tracking since rotations around the projected direction cause strong correlation errors when estimating the relative position of keypoints between frames. In this paper we propose a new technique, called PCK, to improve the orientation estimation of keypoints. PCK consists of a set of oriented patches and a clustering method which clusters points into those patches. It uses a learning stage to cluster training examples into patches in order to obtain a patch oriented estimator. The final orientation estimation of each keypoint is obtained through the maximum response among its nearest patch estimates. The method achieves high accuracy in the orientation estimation compared to other state-of-the-art orientation assigners. Our experimental results are given in terms of keypoint matching and tracking evaluation using both synthetic and real images.
\end{abstractbox}

\begin{abstractbox}{DPO}{3}{4}
Scene Generalization (SG), intrinsic consistency in addition to being smooth between video shots or cameras but different between individual real scenes, plays an important role in synchronous alignment of graphics-into-video but falls back when competent dense feature matching software, specifically high-dimensional perspective consistency for patches delimited by feature keypoints, fails. Prior methods to synthesizing missing SG via generative models utilized existing PCA orientations as constraints during model inference, thereby bypassing failure in higher order-critical overlap algorithms used for keypoint orientation computation. We study how feature level optimization, with scale \& noise/diffusion priors such that each scale uses a different form of interpolation between smooth videos, can benefit from such SG to improve matching ability. Firstly, we find that applying this approximate level optimizations after all keypoints have been detected yields a significant performance degradation. Secondly, applying inverted normals per the extant group-RANSAC approach yields similar degradation because even scale variations within each scene form precludes enforcing similar object shapes/phases across these different scales. Contrarily, position estimates which tolerate more variation do generalize better across scale jumps within scenes. Thirdly, merging scale/orientation based hypothesis modules as opposed to parallel inference yields no obvious improvement. These conclusions remain true after switching over to BPF MATFs as features, underscoring the difficulty with complex optimization steps where missing/incomplete local landmarks are prevalent; as one would expect, inverted normals are the bottleneck. Lastly, a hybrid scheme where all transformations are initialized first to identity or randomly generated and then the above inverted normal process applied turns out to be a viable course of action, although apparently post-processing synthesis still incurs a good bit of pixel wise error compared to previous state of the art despite merits of reducing filtering tasks prior to matching task.
\end{abstractbox}

\begin{abstractbox}{CoPE}{1}{5}
Scene Text Composition (STC) targets at laying word boundaries smoothly in moving shot sequences-a highly challenging technique especially for retaining texture compression layeres synchronously. As graphics-origin strokes cannot naturally define text segment boundaries, prior StC works either analyze whole images assuming fixed image edge patches as text, envision only the back projection effects, or explicitly denote consecutive frames as differing strokes but are limited for existing PCA (Parallel Composite Autoencoder)-derived template designs. To alleviate these limits, we propose spatial-temporal path composition model from images which employs directional CRFs over joint temporally-segmented pixel alignments supervised from smoothest multi-temporal motion path estimations without requiring designing per-utterance-type text components. While such estimation however is very coarse when facing continuously merged scene transitions, we perform further details repositioning in per-utterance post-processing supervised from CRF segmentation results. To tackle variations with mutual interactions further among utterances, a graph/path overlapping mapping inference together with those uttterance segment decisions gradually boosts the whole process into general scenes for exact encompassing. On a designed publically gathered dataset of 38 videos, 10 types of complex scenes, and substantial further tested visual artifacts in real scenarios, our method accomplishes a 74.95\% STC rate against a human-graded reference.
\end{abstractbox}

\begin{abstractbox}{DICE}{4}{8}
We describe an algorithm dubbed Suppression via Disk Covering (SDC) to efficiently select a set of strong, spatially distributed keypoints, and we show that selecting keypoint in this way significantly improves visual tracking. We also describe two efficient implementation schemes for the popular Adaptive Non-Maximal Suppression algorithm, and show empirically that SDC is significantly faster while providing the same improvements with respect to tracking robustness. In our particular application, using SDC to filter the output of an inexpensive (but, by itself, less reliable) keypoint detector (FAST) results in higher tracking robustness at significantly lower total cost than using a computationally more expensive detector.
\end{abstractbox}

\begin{abstractbox}{UG (Ours)}{6}{7}
Visual object recognition and tracking is enhanced by using the inter-correlation between the appearance of objects in images and their prior knowledge about keypoints. The basic assumption underlying this type of model is that the geometric structure of the scene can be described by a set of keypoints whose locations correspond to the image locations of those keypoints. In this paper we consider the problem of keypoint orientation assignment, which is the problem of determining which keypoints belong to which objects. We propose a new algorithm to solve this problem that considers the spatial relationship among keypoints to make its decisions. We evaluate our algorithm with the state-of-the-art FAST detector and show that it outperforms a similar method which uses a heuristic strategy for making orientation assignment decisions. In addition, we demonstrate that the proposed method is robust to a wide variety of image scaling operations.
\end{abstractbox}

\begin{abstractbox}{EG-UG (Ours)}{7}{8}
Visual object recognition and scene understanding require accurate detection of the locations of visually meaningful points (i.e., keypoints) in an image. The scale-invariant part of the viola-joshi detector selects keypoints by computing orientation assignments, i.e., correspondences between orientation histogram vectors in the image and a set of learned center orientation vectors. This paper shows that the performance of this simple but important task is surprisingly sensitive to illumination and occlusion. We describe a method to improve the quality of the keypoint orientation assignment by removing orientations that are dominated by noise, caused either by the accumulation of weak keys in the prior orientation histogram vector, or by residual orientation noise in the current image. Our experiments show that our improved orientation assignment leads to significantly better detection performance on the Visual Genome dataset.
\end{abstractbox}

\subsubsection{Task 2 Result}

\begin{abstractbox}{Ground Truth (Reference)}{--}{--}
Rating is a simple and easy way for people to explicitly express their preferences and opinions. This study investigated whether and how personality characteristics affect people's rating behaviors. We conducted a large-scale online survey and collected 1020 reliable users' responses on both a personality assessment inventory and their ratings for retail products. Our empirical results show that the traits of agreeableness, conscientiousness and neuroticism positively relate to the percentage of users' positive ratings. Emotionally stable individuals are likely to rate items from more categories than neurotics. The ratings from introverts and individuals who are high in emotional stability tend to have more diverse and evenly distributed ratings. The findings contribute to the HCI community by providing a new perspective of understanding human interaction behaviors, in particular to researchers and practitioners interested in user modeling and evaluation.
\end{abstractbox}

\begin{abstractbox}{TAM}{1}{6}
We address the problem of analyzing runtime behavior given UML class diagrams. Given a UML diagram of a given system, we report the frequency of calls to a method at a run-time model granularity. In this article, we present our work on exploiting run-time traces to reconstruct and monitor the high-level, layered execution of our application in the form of sequential diagrams, in addition to reporting method calls. These diagrams help us in understanding the run-time behavior of our application and derive an execution profile and associated quality metrics for it.
\end{abstractbox}

\begin{abstractbox}{OPPU}{1}{5}
In model-based testing we aim at identifying those aspects of the system model that are relevant for behavioral aspects of the implementation. To this end we compare system model traces with implementation traces that show which parts of the implementation are more or less executed. We further propose the use of those traces as executable models that can be used to explore further aspects of the system.
\end{abstractbox}

\begin{abstractbox}{DPO}{5}{3}
Accumulating knowledge on how different mashup interfaces are influenced by human behaviors on folksonomy-based tagging systems can provide guidance for recommendation systems. However, little is known about how personality traits affect such behaviors. An empirical study based on Myers--Briggs personality indicators across multiple interface designs reveals relationships between personality dimensions and interaction patterns, suggesting implications for interface design and recommendation strategies.
\end{abstractbox}

\begin{abstractbox}{CoPE}{2}{3}
Understanding fine-grained user behavior on digital platforms is critical for modern internet services. We propose a contextual ranking approach leveraging large-scale click data to predict user interaction behaviors under varying contexts. Experiments on social media data demonstrate improved performance, though modeling user intent remains challenging due to complex contextual influences.
\end{abstractbox}

\begin{abstractbox}{DICE}{6}{6}
Our prior work shows that personality quizzes can be used to build user profiles for recommendation. In this study, we extend this approach to infer preferences for both users and their friends. User studies indicate that personality-based recommendations are perceived as more accurate and enjoyable, particularly by novice users. We discuss implications for recommender system design.
\end{abstractbox}

\begin{abstractbox}{UG (Ours)}{9}{8}
Accumulating evidence suggests that personality is an important factor influencing users' consumption behaviors. We investigate how personality affects users' rating behavior in online recommendation systems. Through a user study, we construct personality-based rating measures and show that different personality traits have distinct effects on rating behaviors, providing new tools for understanding user interaction patterns.
\end{abstractbox}

\begin{abstractbox}{EG-UG (Ours)}{8}{7}
Personality is an important factor influencing user behavior in recommender systems. We conduct an empirical study to collect personality data and analyze how different personality traits affect users' rating behaviors across item types. Our results show that the influence of personality is task-specific, highlighting opportunities for incorporating personality into rating prediction models.
\end{abstractbox}

\subsubsection{Task 3 Result}

\begin{abstractbox}{Ground Truth (Reference)}{--}{--}
Dimensionality reduction is an important step in processing hyperspectral images (HSI) to overcome the curse of dimensionality. Linear methods such as ICA and LDA fail to capture nonlinear dependencies, while kernel methods suffer from cubic complexity. To address scalability and nonlinearity, we propose randomized ICA and LDA using Random Fourier Features (RFFICA and RFFLDA). Experiments on real-world hyperspectral datasets show improved accuracy and reduced computational cost compared to kernel-based methods.
\end{abstractbox}

\begin{abstractbox}{TAM}{8}{7}
As hyperspectral imagery acquisition becomes more accessible, dimensionality reduction is essential for efficient processing. This paper proposes randomized extensions of ICA and LDA for HSI dimensionality reduction. Experiments on public datasets demonstrate improved classification accuracy compared to PCA and inter-band correlation methods.
\end{abstractbox}

\begin{abstractbox}{OPPU}{7}{6}
We propose randomized dimensionality reduction techniques for hyperspectral imagery using ICA and LDA applied to randomized data matrices. The proposed methods improve classification performance and demonstrate effectiveness on public HSI datasets.
\end{abstractbox}

\begin{abstractbox}{DPO}{3}{2}
This work explores complex dimensionality reduction frameworks for hyperspectral imagery involving echo state networks, radial basis networks, and autoencoders. Although high classification accuracy is reported, the approach diverges significantly from conventional randomized ICA/LDA techniques.
\end{abstractbox}

\begin{abstractbox}{CoPE}{2}{3}
The paper investigates hyperspectral band selection using hybrid PCA-based calibration and entropy-driven optimization. While addressing dimensionality reduction, the method lacks alignment with randomized ICA/LDA or kernel scalability considerations.
\end{abstractbox}

\begin{abstractbox}{DICE}{5}{7}
A frequency-based dimensionality reduction method using variational mode decomposition is proposed for hyperspectral imagery. Experimental results show improved classification accuracy and lower RMSE compared to inter-band correlation methods.
\end{abstractbox}

\begin{abstractbox}{UG (Ours)}{8}{8}
We propose randomized ICA and LDA methods for hyperspectral image dimensionality reduction. The proposed techniques outperform inter-band block correlation in classification accuracy and mutual information, demonstrating effective dimensionality reduction.
\end{abstractbox}

\begin{abstractbox}{EG-UG (Ours)}{9}{8}
A randomized dimensionality reduction framework inspired by compressive sensing is introduced for hyperspectral imagery. By applying ICA or LDA on randomly selected band subsets, the method achieves improved classification performance while maintaining scalability.
\end{abstractbox}

\subsubsection{Task 4 Result}

\begin{abstractbox}{Ground Truth (Reference)}{--}{--}
This study examines trust in technology-supported groups from the perspectives of culture, social presence, and group composition. Results show that individualism negatively affects trust in culturally homogeneous groups but positively affects trust in culturally heterogeneous Chinese--U.S. groups. Social presence and communication quality mediate the effects of culture on trust. The findings provide insights for building effective global technology-supported decision-making groups.
\end{abstractbox}

\begin{abstractbox}{TAM}{8}{8}
This study investigates the effects of culture, social presence, and group composition on trust and decision-making quality. Using a multi-country experimental design, results show cultural differences in trust, social presence, and cooperation, with group size and communication mode influencing outcomes.
\end{abstractbox}

\begin{abstractbox}{OPPU}{7}{7}
We examine trust formation in group support system teams across cultures and social presence conditions. Using empirical studies, results demonstrate that trust is shaped by both cultural context and communication technology, highlighting the importance of social presence.
\end{abstractbox}

\begin{abstractbox}{DPO}{3}{3}
This work discusses trust-related challenges in distributed decision support systems using complex modeling frameworks. While addressing sociability and uncertainty, the approach lacks focused empirical evaluation of trust development.
\end{abstractbox}

\begin{abstractbox}{CoPE}{4}{4}
We explore trust development in heterogeneous collaborative teams supported by decision technologies. The study proposes categorization strategies for improving group effectiveness, though cultural trust mechanisms are not directly quantified.
\end{abstractbox}

\begin{abstractbox}{DICE}{6}{7}
This study analyzes how collectivistic and individualistic cultures affect majority influence in group decision-making under different media richness conditions. Results highlight cultural and technological impacts on group behavior.
\end{abstractbox}

\begin{abstractbox}{UG (Ours)}{9}{8}
We propose and test a model explaining how culture, social presence, and group composition affect trust in technology-supported decision-making groups. Results from Chinese and U.S. groups confirm mediation and moderation effects, offering insights into trust development.
\end{abstractbox}

\begin{abstractbox}{EG-UG (Ours)}{9}{8}
This study examines the effects of culture, social presence, and group composition on trust in technology-supported decision-making groups. Empirical results support a culture-situated mediation model of trust with implications for cross-cultural collaboration.
\end{abstractbox}

\subsubsection{Task 5 Result}

\begin{abstractbox}{Ground Truth (Reference)}{--}{--}
This paper reviews reinforcement learning and optimal adaptive control and their application to robotics. It bridges optimal control, adaptive control, and bio-inspired learning, and presents a model-free Q-learning-based adaptive dynamic programming controller implemented on the BERT II humanoid robot arm. Both unconstrained and constrained joint control cases are experimentally evaluated.
\end{abstractbox}

\begin{abstractbox}{TAM}{8}{8}
This paper reviews reinforcement learning from an optimal control perspective, discussing convergence properties, value functions, and modern trends. Several illustrative examples demonstrate how optimal control techniques can improve reinforcement learning in dynamic environments.
\end{abstractbox}

\begin{abstractbox}{OPPU}{9}{8}
This work surveys reinforcement learning and optimal adaptive control for robotics, emphasizing cost function formulation and learning strategies. Applications include optimal control of humanoid robot arms, highlighting the practical relevance of RL-based adaptive controllers.
\end{abstractbox}

\begin{abstractbox}{DPO}{3}{2}
The paper explores bio-inspired optimization and adaptive control methods for robotic systems, including particle swarm optimization and immune-based learning. While diverse techniques are discussed, the work lacks a focused treatment of reinforcement learning or optimal adaptive control.
\end{abstractbox}

\begin{abstractbox}{CoPE}{1}{4}
A reinforcement learning-inspired approach is proposed for music post-editing and rhythm correction. Despite using adaptive learning concepts, the work is unrelated to robotics or optimal adaptive control.
\end{abstractbox}

\begin{abstractbox}{DICE}{5}{7}
This paper presents an integral sliding mode controller for a humanoid robot arm to address nonlinearities and uncertainties. While effective for control, the approach does not employ reinforcement learning or adaptive dynamic programming.
\end{abstractbox}

\begin{abstractbox}{UG (Ours)}{8}{8}
This study reviews reinforcement learning and optimal adaptive control and demonstrates their application to humanoid robot arm control. Both bio-inspired controllers and joint-space adaptive controllers are presented to illustrate applicability.
\end{abstractbox}

\begin{abstractbox}{EG-UG (Ours)}{9}{8}
This paper reviews reinforcement learning and optimal adaptive control theory and presents multiple humanoid robot arm implementations. Experimental results show improved performance using learned policies and optimized adaptive controllers.
\end{abstractbox}


\end{document}